\pdfoutput=1

\documentclass[11pt]{article}

\usepackage[]{ACL2023}

\usepackage{times}
\usepackage{latexsym}

\usepackage[T1]{fontenc}

\usepackage[utf8]{inputenc}

\usepackage{microtype}

\usepackage{booktabs}
\usepackage{graphicx}
\usepackage{amsmath,amssymb,amsfonts}
\usepackage[titlenumbered,ruled]{algorithm2e}
\usepackage{csquotes}
\usepackage{pifont}
\usepackage{bbm}
\usepackage[noabbrev]{cleveref}
\usepackage{caption}
\usepackage{subcaption}
\usepackage{tikz}
\usepackage{array}
\newcolumntype{?}{!{\vrule width 1pt}}

\usepackage{inconsolata}

\newcommand{\nospacetext}[1]{\makebox[0pt][l]{#1}}
\newcolumntype{R}[1]{>{\raggedleft\let\newline\\\arraybackslash\hspace{0pt}}m{#1}}
\newcolumntype{L}[1]{>{\raggedright\let\newline\\\arraybackslash\hspace{0pt}}m{#1}}

\newcommand{\auc}{$\Delta\mathrm{AUC}$}
\newcommand{\auccut}{$\Delta\mathrm{AUC}_{10}$}
\newcommand{\pathdifference}{ASM}
\newcommand{\ameasure}{\pathdifference}
\newcommand{\bmeasure}{ADM}

\title{On Dataset Transferability in Active Learning for Transformers}

\author{Fran Jeleni{\'{c}} \quad Josip Juki{\'{c}}  \quad Nina Drobac \quad Jan {\v{S}}najder\\
University of Zagreb, Faculty of Electrical Engineering and Computing, Croatia\\
Text Analysis and Knowledge Engineering Lab\\
\tt \{fran.jelenic, josip.jukic, nina.drobac, jan.snajder\}@fer.hr
} 

\begin{document}
\maketitle

\begin{abstract}

Active learning (AL) aims to reduce labeling costs by querying the examples most beneficial for model learning. While the effectiveness of AL for fine-tuning transformer-based pre-trained language models (PLMs) has been demonstrated, it is less clear to what extent the AL gains obtained with one model transfer to others. We consider the problem of transferability of actively acquired datasets in text classification and investigate whether AL gains persist when a dataset built using AL coupled with a specific PLM is used to train a different PLM. We link the AL dataset transferability to the similarity of instances queried by the different PLMs and show that AL methods with similar acquisition sequences produce highly transferable datasets regardless of the models used. Additionally, we show that the similarity of acquisition sequences is influenced more by the choice of the AL method than the choice of the model.





\end{abstract}

\section{Introduction}


Pre-trained language models (PLMs) -- large over-parameterized models based on the transformer architecture \cite{vaswani-2017-attention} and trained on large corpora -- are the leading paradigm in modern NLP, yielding state-of-the-art results on a wide range of NLP tasks.
However, large models require large amounts of data.
\emph{Active learning} \cite[\textbf{AL};][]{settles-2009-active} addresses the data bottleneck problem by improving data labeling efficiency. It employs human-in-the-loop labeling with the model iteratively selecting data points most informative for labeling.
Recent work has demonstrated the effectiveness of AL for fine-tuning PLMs \cite{dor-2020-active,griesshaber-2020-fine, margatina-2022-importance,yuan-2020-cold, shelmanov-2021-active}.
While AL may considerably reduce model development costs, it also potentially limits the scope of use of the actively acquired datasets. Since data sampling in AL is guided by the inductive bias of the acquisition model, the dataset will typically not represent the original population's distribution \cite{attenberg-2011-inactive}.
This is troublesome if one wishes to use the actively acquired dataset to train a different model (\textit{consumer model}) from the one used for AL (\textit{acquisition model}). If the two models' inductive biases differ, the AL gains can cancel or even revert: the consumer model may perform worse when trained on the actively acquired dataset than on a randomly sampled one. However, the robustness of the actively acquired dataset to the choice of the consumer model is obviously highly desirable, as the acquisition model may become unavailable or dated. The latter is common in NLP, where new and better models are being developed faster than new datasets. However, 
most AL studies use the same acquisition and consumer models, and dataset transferability is seldom mentioned in AL literature. A notable exception is the work of \citet{lowell-2018-practical}, who showed the unreliability of dataset transfer on standard NLP tasks.



In this work, we examine the problem of AL dataset transferability for transformer-based PLMs and conduct a preliminary empirical study on text classification datasets. 
We first probe whether AL gains persist between different transformer-based PLMs, considering several AL methods and datasets. Observing that on most datasets, the transfer works in some cases but fails in others, we investigate the mechanisms underlying transferability. We hypothesize a link between AL dataset transferability and how the acquisition and consumer models sample instances. To probe this, we introduce \emph{acquisition sequence mismatch} (\pathdifference{}) to characterize to what extent the two models differ in how they sample instances throughout AL iterations. We investigate how \pathdifference{} affects dataset transferability and how \pathdifference{} is affected by other AL variables. 
We show that, while it is generally reasonable to transfer actively acquired datasets between transformer-based PLMs, AL methods that retain low \pathdifference{} produce more transferable datasets. We also show that the choice of the AL method affects \pathdifference{} more than the choice of models. 

To summarize our contributions: we (1) conduct an empirical study on the transferability of actively acquired datasets between transformer-based PLMs, (2) propose a measure to quantify the mismatch in the acquisition sequences of AL models and link this to dataset transferability, and (3) analyze what design choices affect this mismatch. We provide code for the experiments\footnote{\url{https://github.com/fjelenic/al-transfer}} with the hope that our results will encourage NLP practitioners to use AL when fine-tuning PLMs and motivate further research into the AL dataset's transferability.

\section{Related Work}


Although AL has been extensively studied for shallow and standard neural models (without pre-training), research on combining AL and PLMs lags behind. The initial studies showed promise, with AL methods outperforming random sampling for text classification \cite{dor-2020-active, griesshaber-2020-fine}. The field is gradually gaining traction with studies demonstrating AL effectiveness even with simple uncertainty-based methods \cite{gonsior-2022-softmax, schroder-2022-revisiting}. Moreover, PLMs open up new possibilities, such as complementing AL with model adaptation using unlabeled data \cite{yuan-2020-cold, margatina-2022-importance}.

While there is much research on AL for standard scenarios where the acquisition and consumer models are the same, there is little research on AL dataset transfer. \citet{prabhu2019sampling} demonstrated that combining uncertainty AL strategies with deep models produces sampled datasets with good sampling properties that have a large overlap with support vectors of SVM trained on the entire dataset. Likewise, \citet{farquhar-2021-statistical} showed that deep neural models benefit from the sample bias induced by the acquisition model (the opposite is true for shallow models). However, the jury is still out on the effects of sample bias on the consumer model. The most prominent empirical study on AL transfer with neural models \cite{lowell-2018-practical} predates PLMs.
\citet{tsvigun2022towards} focused on alleviating the effects of acquisition-consumer mismatch in PLMs by using lightweight distilled models for acquisition and larger versions of the models as consumer models. Even though the study focuses on improving the transferability of actively acquired datasets, the reasons behind the successful transfer are yet to be explored.
An older study of AL dataset transferability for text classification and shallow models by \citet{tomanek-morik-2011-inspecting} showed that transfer works in most cases but that neither sample nor model similarity explains transferability. Our study explores these characteristics for acquisition-consumer pairings of different PLMs.

\section{Experimental Setup}

Our study used four datasets, three models, and three AL methods (cf.~\Cref{sec:appendix:exp-design} for details). The datasets we used are Subjectivity \cite[\textbf{\textsc{subj};}][]{Pang+Lee:04a}, CoLA \cite[\textbf{\textsc{cola};}][]{warstadt-2018-neural}, AG-News \cite[\textbf{\textsc{agn};}][]{zhang-2015-character}, and TREC \cite[\textbf{\textsc{trec};}][]{li-roth-2002-learning}). The three transformer models we used are BERT \cite{devlin-2018-bert}, RoBERTa \cite{liu-02019-roberta}, and ELECTRA \cite{clark-2020-electra}. The AL methods we considered are entropy \cite[\textbf{\textsc{ent};}][]{settles-2009-active}, core-set \cite[\textbf{\textsc{cs};}][]{sener-2017-active}, and BADGE \cite[\textbf{\textsc{ba};}][]{ash-2019-deep}). This gives 108 AL configurations (72 transfer and 36 no-transfer configurations). Furthermore, we ran each configuration with 20 different warm-start sets to account for stochasticity. The AL acquisition was simulated until the budget of 1500 labeled data points was exhausted (model performance for all datasets reached a plateau), labeling 50 data points per step.

We assessed dataset transferability using the difference in the area under the $F_1$ curve of the model trained on the actively acquired dataset and the same model trained on a randomly sampled dataset (\auc{}).
We deem the AL dataset transfer successful if \auc{} is not significantly less than zero and unsuccessful otherwise. We chose \auc{} to make the notion of transferability independent of when the AL acquisition terminates. 
On the other hand, as terminating the AL after acquiring too few labeled data is unrealistic, we also report \auccut{}, which is \auc{} calculated with an offset of 10 iterations (500 labeled instances) of the AL loop. Comparing \auccut{} to \auc{} provides insights into how transferability changes through time.

\section{Results}


\subsection{Dataset transferability}
\label{sec:transferability}

We grouped the 108 AL configurations into three groups based on the sign of the mean \auc{} value and the p-value of the difference between AUC scores of transfer and random sampling:\footnote{We used either the paired t-test or Wilcoxon signed-rank test, depending on the results of Lilliefors’ test for normality.} negative (\auc{} $<$ 0 and p$<$.05), neutral (p$\geq$.05), and positive (\auc{} $\geq$ 0 and p$<$.05) transfer.
The no-transfer AL configurations (where the acquisition and consumer models are the same) are generally successful (25 positive, 9 neutral, and 2 negative configurations as per \auc{}; 33 positive, 2 neutral, and 1 negative configuration as per  \auccut{}).  
The grouping of the remaining 72 configurations with AL dataset transfer is given in \Cref{tab:groups}. We observe that the dataset, the acquisition-consumer model pairing, and the AL method all affect transfer success.

\begin{table}[t]
\small
\centering
\begin{tabular}{l|rrrc|rrrc|r}
\toprule
& $\Delta$\nospacetext{$^-$} & $\Delta$\nospacetext{$^0$} & $\Delta$\nospacetext{$^+$} & & $\Delta$\nospacetext{$_{10}^-$} & $\Delta$\nospacetext{$_{10}^0$} & $\Delta$\nospacetext{$_{10}^+$} & & $\Sigma$ \\
\midrule
\textsc{subj} & $0$ & $0$ & $18$ & & $0$ & $0$ & $18$ & & $18$  \\
\textsc{cola} & $2$ & $8$ & $8$ & & $2$ & $7$ & $9$ & & $18$ \\
\textsc{agn} & $7$ & $4$ & $7$ & & $3$ & $2$ & $13$ & & $18$ \\
\textsc{trec} & $8$ & $3$ & $7$ & & $0$ & $2$ & $16$ & & $18$ \\
\midrule
R$\rightarrow$B & $2$ & $2$ & $8$ & & $0$ & $1$ & $11$ & & $12$ \\
E$\rightarrow$B & $2$ & $2$ & $8$ & & $0$ & $2$ & $10$ & & $12$ \\
B$\rightarrow$R & $2$ & $4$ & $6$ & & $0$ & $1$ & $11$ & & $12$ \\
E$\rightarrow$R & $2$ & $4$ & $6$ & & $1$ & $2$ & $9$ & & $12$ \\
B$\rightarrow$E & $5$ & $1$ & $6$ & & $2$ & $2$ & $8$ & & $12$ \\
R$\rightarrow$E & $4$ & $2$ & $6$ & & $2$ & $3$ & $7$ & & $12$ \\
\midrule
\textsc{ent} & $11$ & $3$ & $10$ & & $3$ & $2$ & $19$ & & $24$ \\
\textsc{cs} & $4$ & $10$ & $10$ & & $2$ & $6$ & $16$ & & $24$ \\
\textsc{ba} & $2$ & $2$ & $20$ & & $0$ & $3$ & $21$ & & $24$ \\
\midrule
 $\Sigma$ & $17$ & $15$ & $40$ & & $5$ & $11$ & $56$ & & \\
\bottomrule
\end{tabular}
\caption{Breakdown of datasets, acquisition$\to$consumer model pairs (denoted by initial letters), and AL methods by transferability: negative ($-$), neutral ($0$), and positive ($+$) transfer. \auc{} is shown as $\Delta$.}
\label{tab:groups}
\end{table}

Evidently, transferability differs across datasets: the transfer is always positive on \textsc{subj} (which is the simplest task we considered in terms of the number of labels, the balance of classes, and the MDL task complexity measure; cf.~\Cref{sec:appendix:exp-design}), while most neutral transfers occur on \textsc{cola}. A more interesting picture emerges from the different acquisition-consumer model pairings and AL methods. Most negative transfers are transfers to ELECTRA, while most neutral transfers are those to RoBERTa (perhaps due to it being optimized for robustness). On the other hand, transfer to BERT is positive in most cases, perhaps because BERT's pre-training regime is most similar to that of the other two models. Among the AL methods, entropy mostly makes the transfer negative, most neutral transfers occur with core-set, and BADGE is the best choice for ensuring positive transferability. However, when looking at the later steps of the AL loop, differences between entropy and BADGE vanish, while the core-set lags slightly behind. Thus, \auc{} tends to increase throughout the AL process, suggesting that increasing the amount of sampled data lowers the risk of unsuccessful transfer (cf.~\Cref{sec:appendix:exp-runs} for additional $F_1$ scores analysis).

\subsection{Acquisition sequence mismatch}

We hypothesize there is a link between dataset transferability and the sequence in which data points are acquired for labeling by AL. In particular, we posit that dataset transferability will be successful when the acquisition sequence of the acquisition model does not differ from what the acquisition sequence of a consumer model would be if that model had access to the original dataset. We introduce the \emph{acquisition sequence mismatch} (\pathdifference{}) to measure the differences in acquisition sequences. To compute the \pathdifference{} between two acquisition sequences, we pair the corresponding batches of the two sequences and average their pairwise differences. To measure the difference between a pair of batches, we take the average of the distances of best-matched examples between the batches. To account for the fact that AL methods may choose numerically different yet semantically similar data points, we measure the similarity of acquired instances in representation space. We use GloVe embeddings  \cite{pennington-2014-glove} as a common representation space independent of the choice of acquisition and consumer models and compute the cosine distance between averaged word embeddings. Lastly, we use the Hungarian algorithm \cite{kuhn-1955-hungarian} to construct a bipartite graph between two batches with distance-weighted edges to find the best-matching examples. Formally, we define \pathdifference{} as follows:
\begin{equation}
\frac{1}{T} \displaystyle\sum_{t=1}^{T} \frac{1}{|B_t|}  \displaystyle\min_{S(B_A^t), S(B_B^t)} \left( \sum_{i=1}^{|B_t|} d(x_A^i,x_B^i) \right) \label{eq:path-difference}
\end{equation}
where $T$ is the length of the sequence (the number of steps of the AL loop), $S(B^t)$ is the set of all of the permutations of instances in the selected batch at step $t$, and $d(x_A^i,x_B^i)$ is the cosine distance between instance representations from sequences $A$ and $B$ for a batch at position $i$ of a given batch permutation. Intuitively, \pathdifference{} assumes that both batches cater to the same informational need of the model, so it calculates how much the instances that should carry out the same role in the batch differ.

\begin{figure}[t!]
\centering
\begin{subfigure}{1.0\linewidth}
  \centering
  \includegraphics[width=\linewidth]{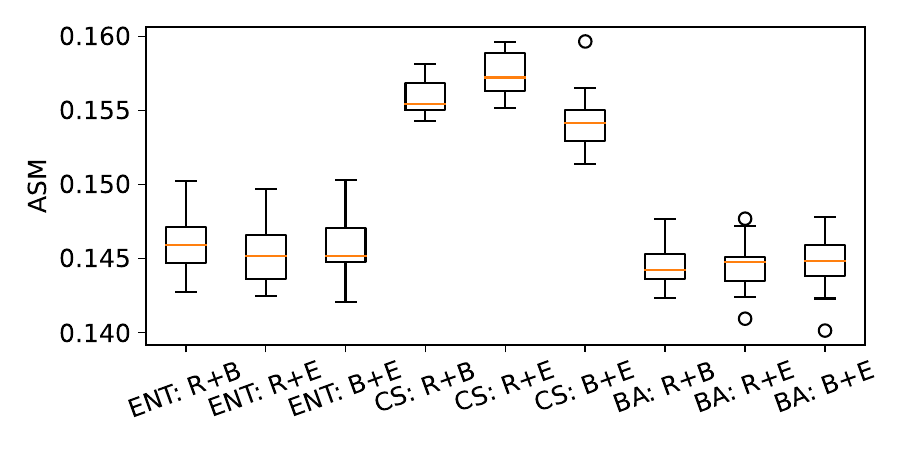}
  \caption{Subjectivity}
\end{subfigure}
\begin{subfigure}{1.0\linewidth}
  \centering
  \includegraphics[width=\linewidth]{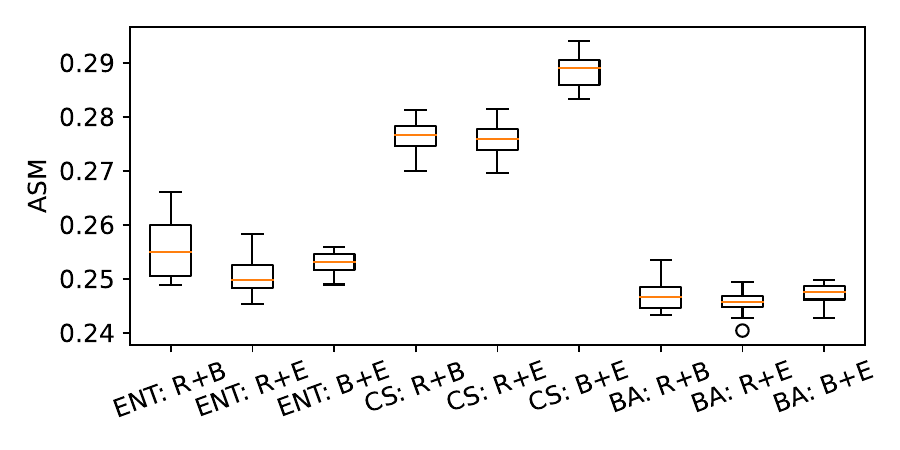}
  \caption{CoLA}
\end{subfigure}
\begin{subfigure}{1.0\linewidth}
  \centering
  \includegraphics[width=\linewidth]{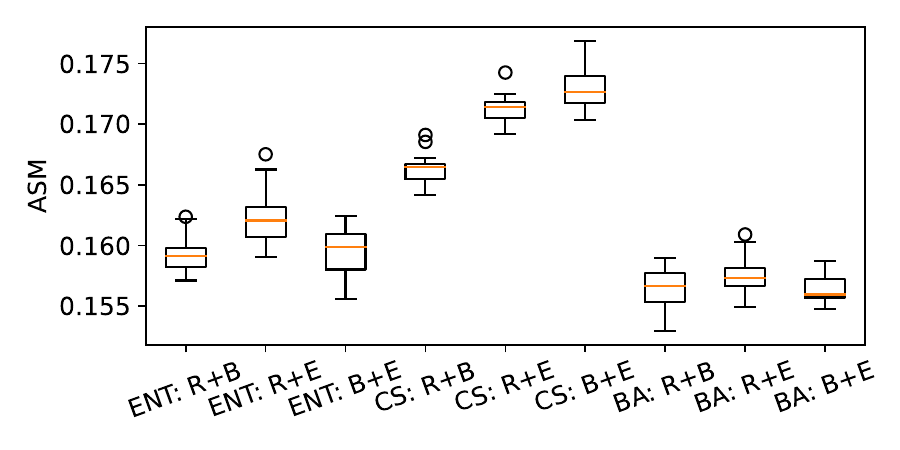}
  \caption{AG-News}
\end{subfigure}
\begin{subfigure}{1.0\linewidth}
  \centering
  \includegraphics[width=\linewidth]{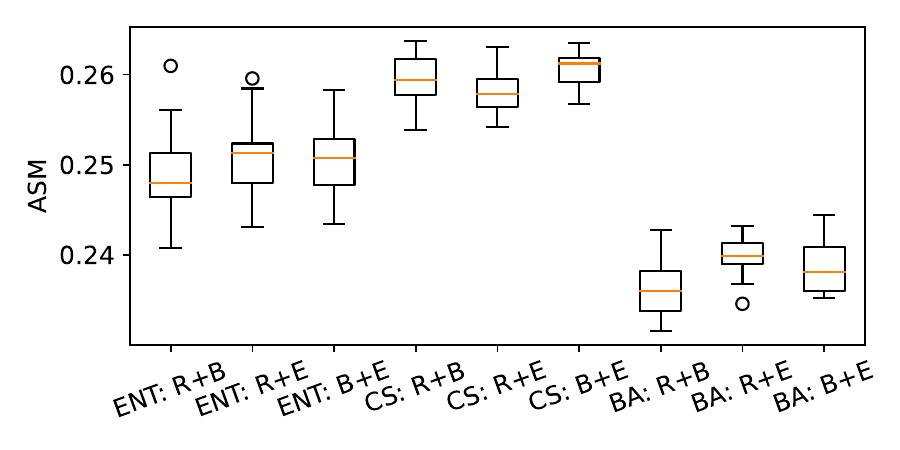}
  \caption{TREC}
\end{subfigure}
\caption{Distributions of \pathdifference{} values for combinations of AL methods and acquisition+consumer model pairs (denoted by initial letters).}
\label{fig:method-vs-model}
\end{figure}

Given a dataset, we hypothesize \pathdifference{} may be affected by both the choice of the models and the choice of the AL method. Figure~\ref{fig:method-vs-model} shows that the distributions of \pathdifference{} values are more alike when grouped by the AL methods than when grouped by the model pairings. 
To verify this observation, we conducted two Kruskal-Wallis H-tests for each dataset: in the first, populations were determined by the AL method, and we concluded that there was a significant difference in \pathdifference{} (p$<$.05); in the second, the populations were determined by the model pairing, and there was no significant difference in \pathdifference{} (p$>$.05). This suggests that the choice of AL method affects \pathdifference{} more than the choice of acquisition-consumer model pairing.

\subsection{Acquisition mismatch analysis}

\begin{figure*}[t!]
\small
\centering
\begin{subfigure}{.24\linewidth}
  \centering
  \includegraphics[width=\linewidth]{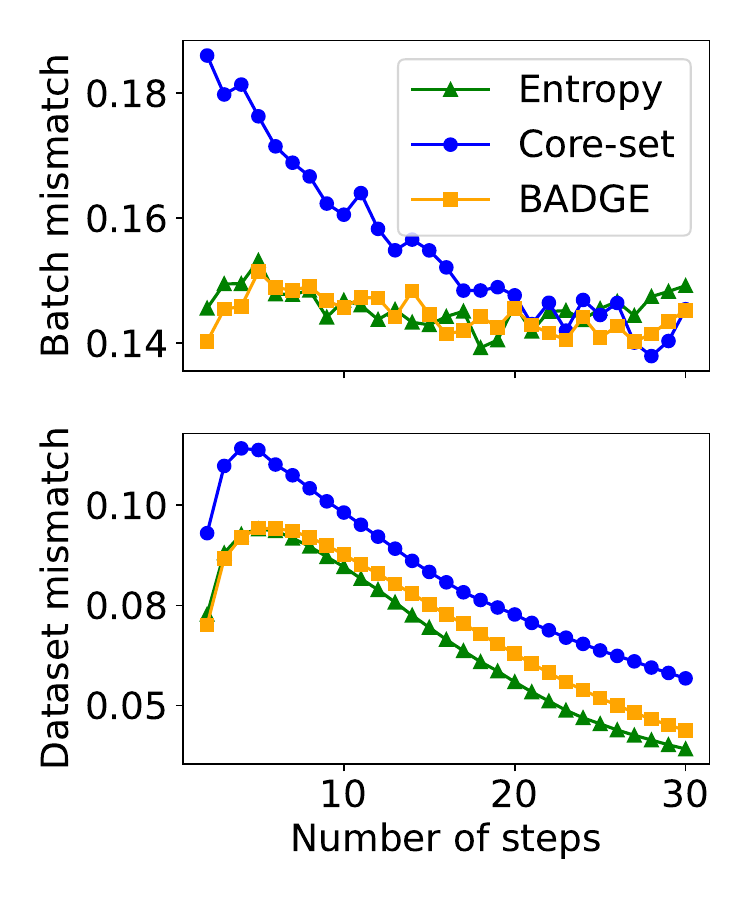}
  \caption{Subjectivity}
\end{subfigure}
\begin{subfigure}{.24\linewidth}
  \centering
  \includegraphics[width=\linewidth]{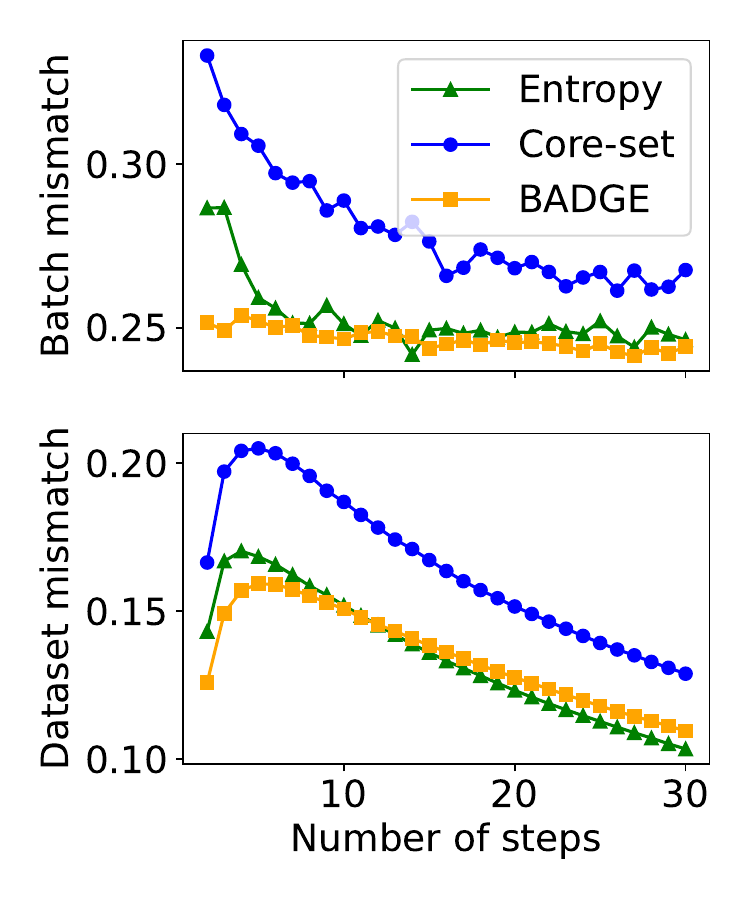}
  \caption{CoLA}
\end{subfigure}
\begin{subfigure}{.24\linewidth}
  \centering
  \includegraphics[width=\linewidth]{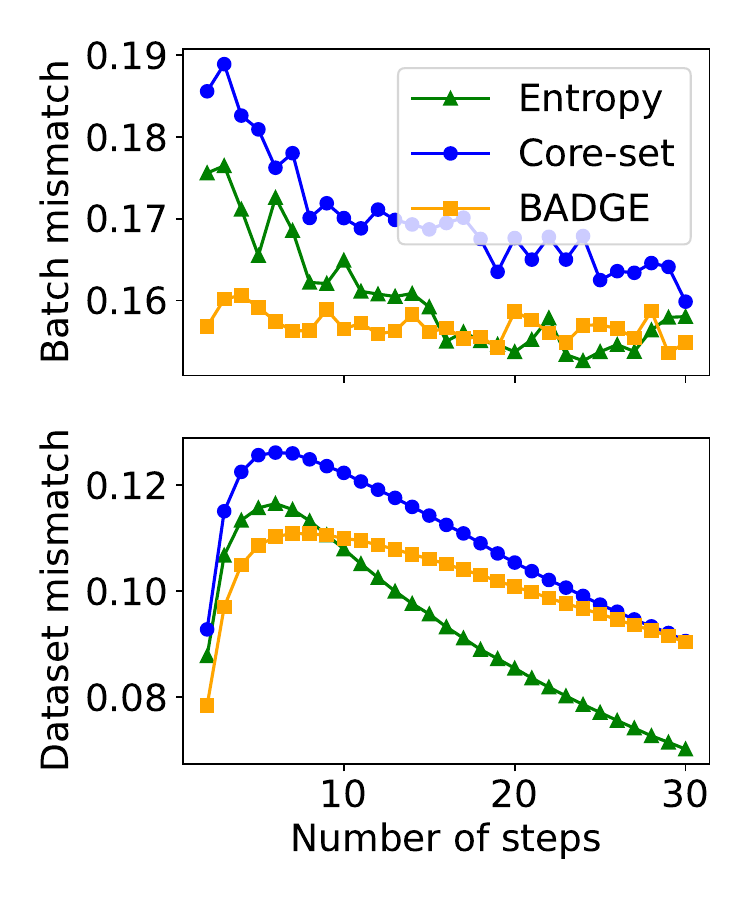}
  \caption{AG-News}
\end{subfigure}
\begin{subfigure}{.24\linewidth}
  \centering
  \includegraphics[width=\linewidth]{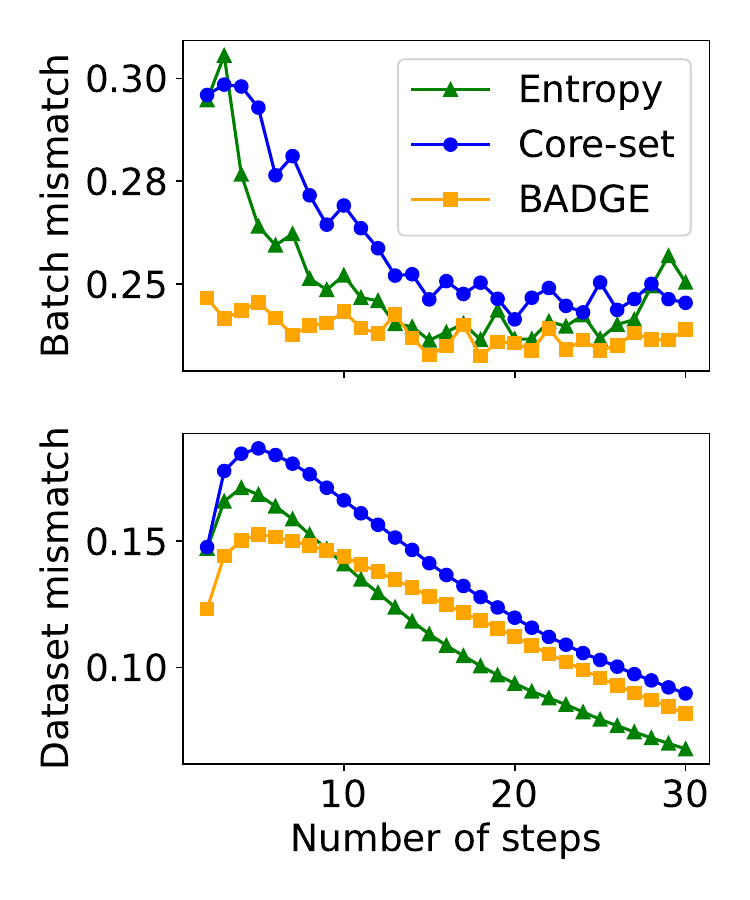}
  \caption{TREC}
\end{subfigure}
\caption{The mismatch between acquired batches (top) and \bmeasure{} at each step of the AL loop (bottom) for different AL methods.}
\label{fig:temporal}
\end{figure*}

We found a statistically significant negative correlation between \auc{} and \ameasure{} for each dataset.\footnote{Spearman correlation coefficients are $-0.11$ for \textsc{subj}, $-0.19$ for \textsc{cola}, $-0.27$ for \textsc{agn}, and $-0.38$ for \textsc{trec}, all significant with p$<$.05.} This supports our hypothesis that the lower the mismatch between acquisition sequences of the two models, the higher the transferability of a dataset from one model to the other.
Besides ASM, we use another measure for analyzing dataset transferability: the difference between the dataset acquired with AL using the acquisition model and the dataset acquired with AL using the consumer model. We call this measure the \emph{acquired dataset mismatch} (\bmeasure{}). Essentially, \bmeasure{} computes the mismatch between samples similarly to ASM but between entire datasets obtained after the last sampling step.



Above we showed that the choice of the AL method affects the \pathdifference{}. \Cref{fig:temporal} shows that BADGE gives smaller \pathdifference{} than the other two methods, whereas core-set gives larger \pathdifference{} than the other two methods.\footnote{Verified using three one-sided Wilcoxon signed-rank tests with p$<$.05 corrected for FWER.} 
However, the intriguing effect emerges when comparing the difference in batches through time and differences in the entire acquired datasets through time. In the early steps, BADGE gives the highest similarity of acquired datasets among the considered methods, which leads to it having the lowest \pathdifference{}.  
However, in later steps, entropy dominates the similarity of acquired datasets.\footnote{Verified using three one-sided Wilcoxon signed-rank tests on \bmeasure{} with p$<$.05 corrected for FWER.} It seems as if entropy acquired similar datasets for different models by taking those models through different sequences of the population distribution. This effect is seen in \Cref{tab:groups}, where entropy is the worst method when using \auc{} to measure transfer success while managing to parry BADGE when using \auccut{}. The difference in transferability between entropy and BADGE completely vanishes when looking at the last step of the AL loop (cf.~Appendix, \Cref{tab:groups-f1}). It is clear that entropy can produce transferable datasets, but it requires more time to do so.  

We speculate that the effect of BADGE having the lowest \pathdifference{} yet entropy achieving the lowest \bmeasure{} could emerge due to the interaction between the AL method and the model's decision boundary. Namely, uncertainty AL methods sample data points on the decision boundary with high overlap with support vectors of the SVM trained on the whole dataset, as pointed out by \citet{prabhu2019sampling}. Since BADGE combines uncertainty and diversity, i.e., it samples data points the model is uncertain about for diverse reasons, it samples along the entire decision boundary at each step, and since decision boundaries of the models are roughly the same, so are the sampled data points. Entropy, on the other hand, relies solely on uncertainty. Due to its greedy nature, entropy tends to sample similar points because if one data point has high uncertainty, data points similar to it are also going to have high uncertainty \cite{zhdanov2019diverse}. This may manifest as sampling local patches of space on the decision boundary. Therefore, entropy may take more time to define the boundary than BADGE because it is forming the boundary from patches of space with the highest uncertainty at a given AL step rather than holistically sampling along the boundary at each step. Since the shape of the decision boundary is more similar between different models than the local interactions along the boundary, entropy has a higher batch mismatch in the early steps. However, once more data is labeled and the boundary becomes stable, both entropy and BADGE start to have a low batch mismatch, as seen in \Cref{fig:temporal}. Since entropy is deterministic and never strays from the decision boundary, it ends up having a lower \bmeasure{} than BADGE. Lastly, we believe that the core-set method has the highest \pathdifference{} and \bmeasure{} because it selects data based on diversity in the model's representation space, which is more model-specific
than the decision boundary. Further exploring the described interaction is a compelling direction for future work.

It may be that AL methods with different acquisition sequences end up acquiring a similar dataset and have high transferability, as in the case of entropy, an uncertainty-based acquisition function. It is also possible that acquired datasets differ between models but that the transfer remains successful because it taps into some other essential aspect of a transferable dataset, as is the case with core-set, a diversity-based acquisition function. However, the best strategy to ensure dataset transferability appears to be a mixture of uncertainty and diversity, as provided by BADGE. This appears to minimize \pathdifference{} between models, making datasets transferable regardless of the number of AL steps.
\section{Conclusion}

We presented an empirical study on the transferability of actively acquired text classification datasets for transformer-based PLMs. Our results indicate no significant risk in transferring datasets, especially for larger amounts of data. We also showed that transfer is largely successful when preserving the sequence and similarity of acquired instances between the models, which is what methods combining uncertainty and diversity acquisition functions seem to do.
Transferability appears to differ considerably across datasets, so future work should examine what dataset characteristics are predictive of transfer success. 


\section*{Limitations}

Our study revealed considerable differences in transferability and other measures we considered across different datasets. Nonetheless, the study focused on the differences in transferability arising from the choice of the models and the AL methods rather than the dataset. To eliminate confounding due to datasets, we grouped the results by datasets and analyzed each group separately. Despite this, the scope of our results is limited by the fact that all datasets used are in English and possibly contain their own biases.

Even though we showed that it could still be useful to transfer actively acquired datasets between transformer-based PLMs, it is important to keep in mind that actively acquired datasets are not representative of the original data distribution due to the sampling bias introduced by active learning.
\section*{Acknowledgments}

This research was supported by the AIDWAS KK.01.2.1.02.0285 grant. We thank the anonymous reviewers for their insightful comments and suggestions.

\bibliography{anthology,references}
\bibliographystyle{acl_natbib}

\clearpage
\appendix

\section{Reproducibility}
\label{sec:appendix:reproduce}
We conducted our experiments on 4× AMD Ryzen Threadripper 3970X 32-Core Processors and 4× NVIDIA GeForce RTX 3090 GPUs with 24GB of RAM, which took roughly one week. We used PyTorch version 1.12.1, Transformers version 4.21.3, and CUDA 11.4.

\section{Experimental design choices}
\label{sec:appendix:exp-design}

\subsection{Datasets}

\begin{table}[]
\small
\centering
\begin{tabular}{lrrrrr}
\toprule
& Train & Test & \# Labels & NLE & MDL\\
\midrule
\textsc{subj} & $8000$ & $2000$ & $2$ & $1.00$ & $0.30$ \\
\textsc{cola} & $8551$ & $1043$ & $2$ & $0.88$ & $1.00$ \\
\textsc{agn} & $20000^*$ & $7600$ & $4$ & $1.00$ & $0.56$ \\
\textsc{trec} & $5452$ & $500$ & $6$ & $0.92$ & $0.34$ \\
\bottomrule
\end{tabular}
\caption{Dataset statistics. We report train and test set sizes, number of labels, normalized label entropy (information entropy of label distribution normalized by the entropy of uniform distribution with the same number of variables), and MDL normalized by the MDL of the dataset with the largest value (\textsc{cola}). Train set of \textsc{agn} was subsampled from the original train set of 120000 instances.}
\label{tab:datasets}
\end{table}

The datasets used in this paper are standard benchmarks in NLP for text classification. We chose these datasets to represent different attributes: the number of labels (binary or multi-class classification) and the balancing of the labels (balanced and imbalanced classes). The diversity of the dataset characteristics can give an insight into the impact of these attributes on dataset transferability. We present dataset statistics in \Cref{tab:datasets}. There we also show \textit{minimum description length} (MDL) \cite{perez-2021-rissanen} of each dataset, which can be interpreted as the complexity of the task.\\
\textbf{Subjectivity:} Movie-review data with reviews labeled as either subjective or objective. This is a balanced dataset with binary labels.\\
\textbf{CoLA:} The Corpus of Linguistic Acceptability is a dataset containing sentences labeled as grammatical or not. This is an imbalanced dataset with binary labels.\\
\textbf{AG-News:} Corpus of news articles annotated by the article's topic (World, Sports, Business, Sci/Tech). The dataset was created by subsampling the corpus to the size of 20,000 examples. This is a balanced dataset with four classes.\\
\textbf{TREC:} The dataset contains questions labeled with the type of subject of the question. This is an imbalanced dataset with six classes.

\subsection{Models}
We picked the models that share the common architecture; they are all transformer-based PLMs but differ in pre-training data and pre-training objectives. This choice of models enables us to analyze the impact of different pre-training design choices on dataset transferability. All models were trained using ADAM optimizer with a learning rate of $2\cdot10^{-5}$ and batch size of 64 for five epochs for both acquisition and evaluation phases.\\
\textbf{BERT:} One of the first and most popular transformer-based pre-trained language models. The model was pre-trained using a generative masked language modeling objective. This model has 12 layers, a hidden state size of 768, and 12 heads with 110M parameters in total.\\
\textbf{RoBERTa:} A model with the same architecture and pre-training objective as BERT but trained on more data and with optimized hyperparameters to make the model more robust. This model has 12 layers, a hidden state size of 768, and 12 heads with 125M parameters in total.\\
\textbf{ELECTRA:} It uses the same architecture and pre-training data as BERT but with discriminative instead of generative pre-training objectives. Instead of masking some tokens in text and having to guess the identity of masked tokens as BERT does, the generative pre-training objective corrupts some tokens by replacing them with plausible alternatives, and then the model has to decide for each token whether it is the original token or the replaced one. This model has 12 layers, a hidden state size of 768, and 12 heads with 110M parameters in total.

\subsection{AL methods}
AL methods used to select the most informative data points are divided into two types of heuristics: uncertainty and diversity. Methods using uncertainty as a heuristic select data based on some measure of the model's uncertainty. The intuition behind the uncertainty methods is that the more uncertain the model is about a data point, the more it can learn from knowing its label. In comparison, diversity-based methods try to represent the input space (which is not always the same as the input population) as accurately as possible with as few data points as possible. AL methods can combine those two heuristics to select a group of data points the model is uncertain about for different reasons.

The choice of the AL methods used in this experiment was motivated by the type of heuristic (uncertainty vs. diversity) they used for sampling. These methods allow us to analyze the impact of the choice of heuristic on the success of dataset transfer in AL.\\
\textbf{Entropy:} An uncertainty-based method that selects data points with maximal information entropy of their posterior class distribution.\\
\textbf{Core-set:} This diversity-based method selects data points that best cover the representation space.\\
\textbf{BADGE:} A method that combines uncertainty and diversity by using $k$-\textsc{means++} algorithm on the would-be gradients of the models' last layer for the data points if their most probable labels were their actual labels.

\section{Experiment runs}
\label{sec:appendix:exp-runs}

\begin{figure*}[t!]
\centering
\includegraphics[width=\linewidth]{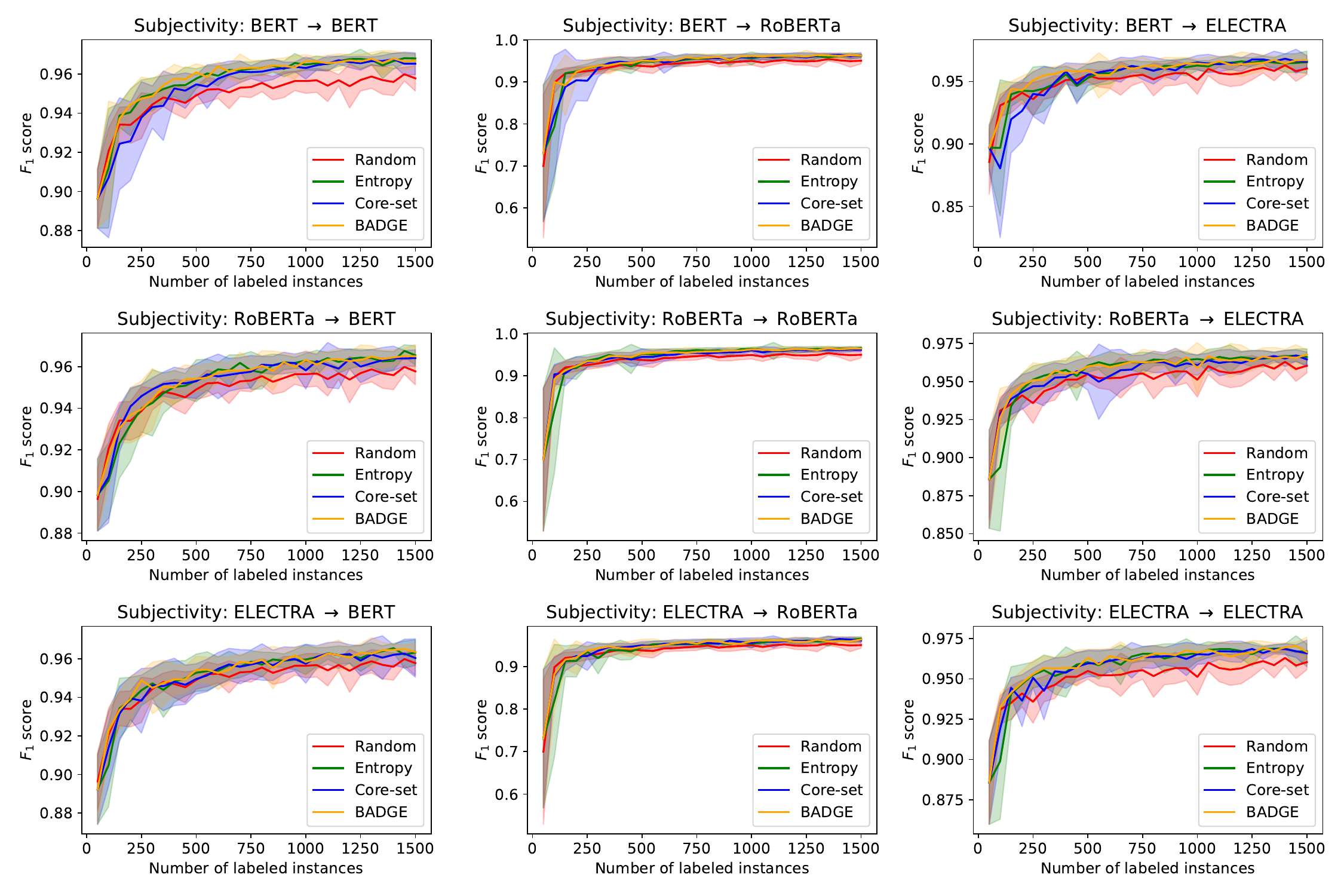}
\caption{Learning curves for the Subjectivity dataset. The figure shows the mean $F_1$ score of 20 runs with confidence intervals of $\pm$ standard deviation.}
\label{fig:al-curves-subj}
\end{figure*}

\begin{figure*}[t!]
\centering
\includegraphics[width=\linewidth]{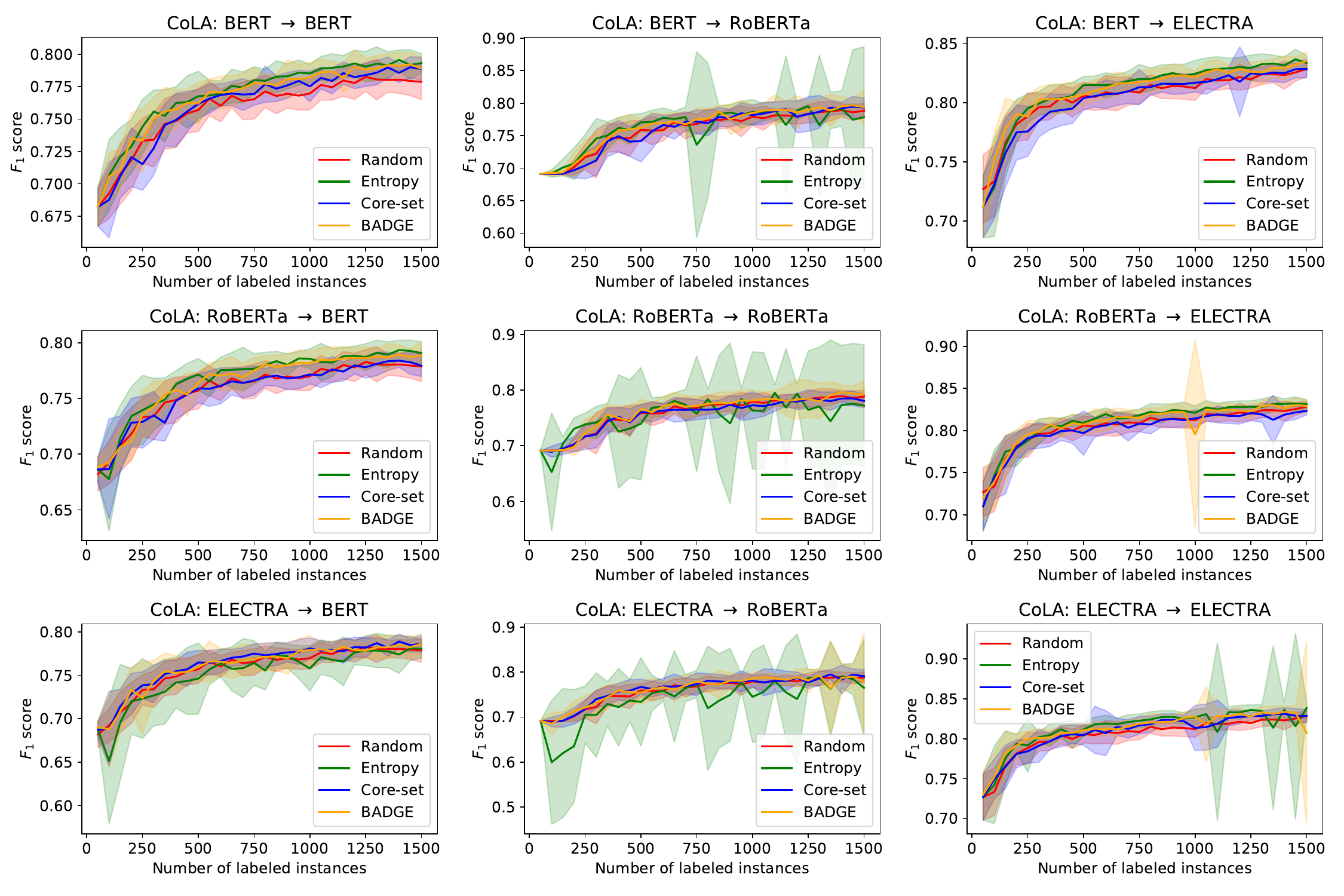}
\caption{Learning curves for the CoLA dataset. The figure shows the mean $F_1$ score of 20 runs with confidence intervals of $\pm$ standard deviation. $F_1$ curves for RoBERTa as consumer model with entropy as AL method have high variance because entropy tends to favor minority class heavily, and the model starts to classify with minority class more often than it should, so the $F_1$ on the test set drastically drops. These drops happen one to two times per seed during the AL loop before the method balances out the labels again.}
\label{fig:al-curves-cola}
\end{figure*}

\begin{figure*}[t!]
\centering
\includegraphics[width=\linewidth]{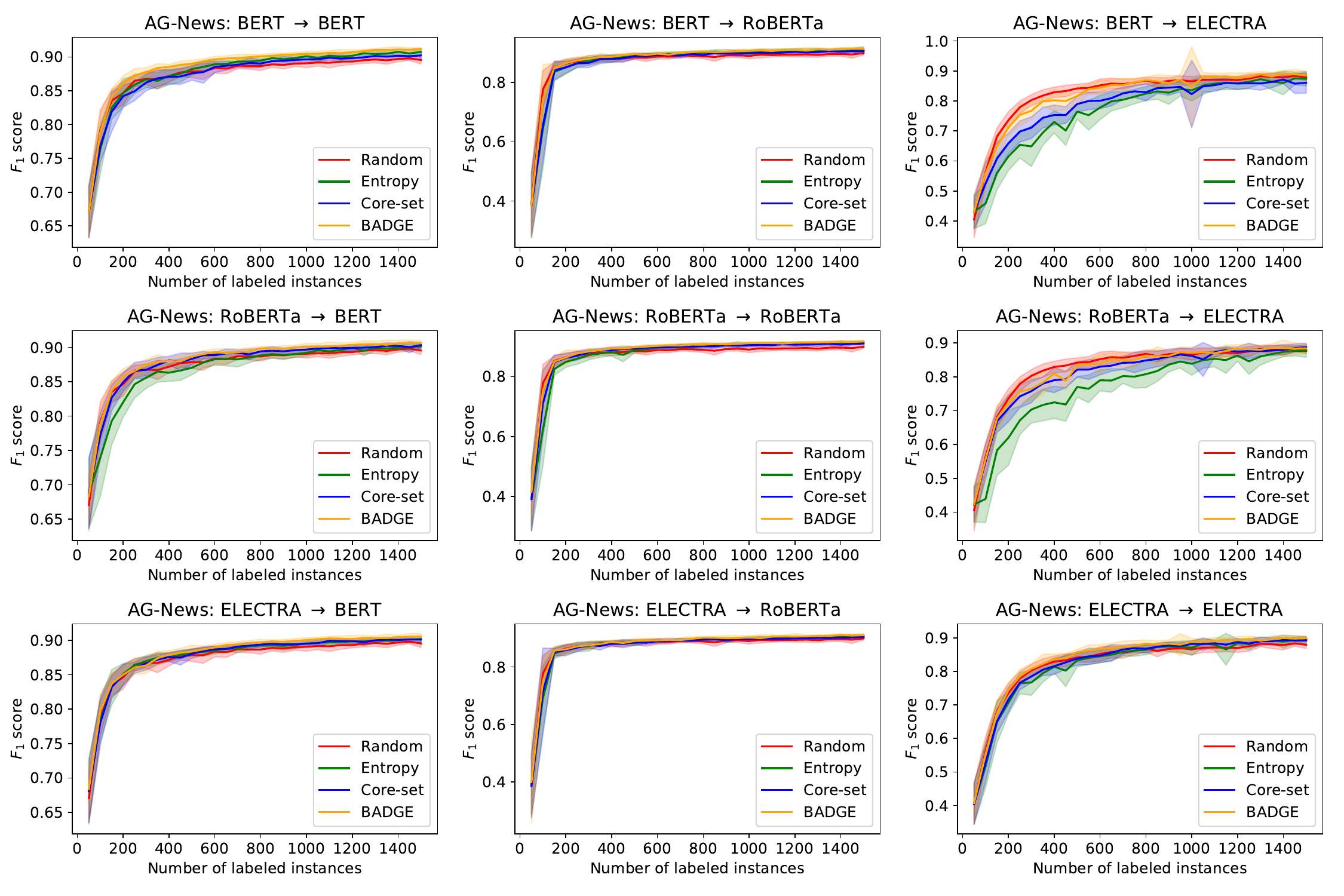}
\caption{Learning curves for the AG-News dataset. The figure shows the mean $F_1$ score of 20 runs with confidence intervals of $\pm$ standard deviation.}
\label{fig:al-curves-agn}
\end{figure*}

\begin{figure*}[t!]
\centering
\includegraphics[width=\linewidth]{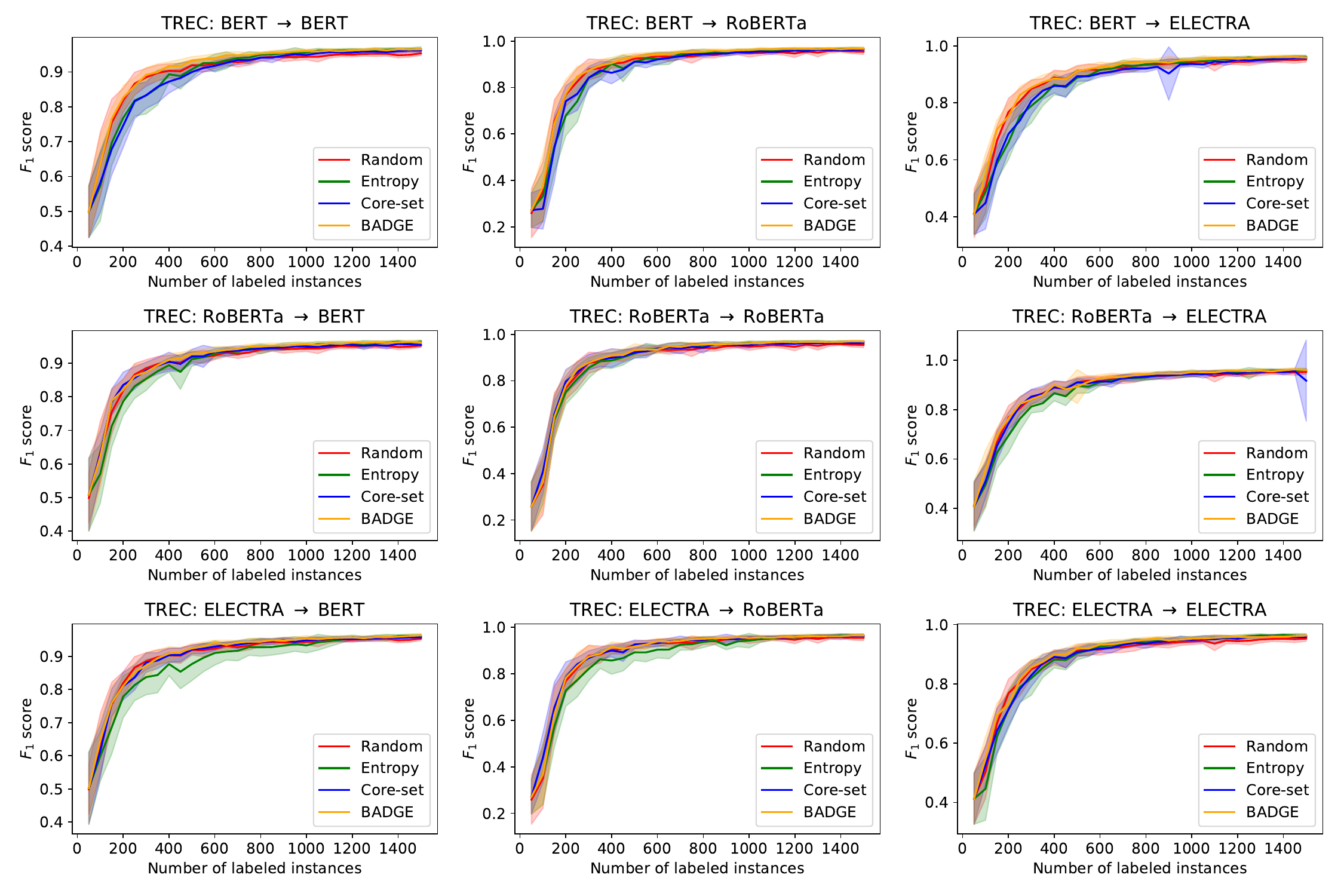}
\caption{Learning curves for the TREC dataset. The figure shows the mean $F_1$ score of 20 runs with confidence intervals of $\pm$ standard deviation.}
\label{fig:al-curves-trec}
\end{figure*}

This section presents more results from our experiment to complement the already presented results. \Cref{tab:groups-f1} shows transferability for different combinations in the fashion of \Cref{tab:groups}. However, instead of measuring transferability with \auc{} this table uses the $F_1$ score at the end of the AL loop (1500 labeled instances). To illustrate the success of regular AL (without the transfer), we present \Cref{tab:groups-regular}. That table shows the same information as \Cref{tab:groups} and \Cref{tab:groups-f1} but for situations where acquisition and consumer models are the same. Lastly, we present the learning curves of all of the runs of the experiment in \Cref{fig:al-curves-subj} for Subjectivity, \Cref{fig:al-curves-cola} for CoLA, \Cref{fig:al-curves-agn} for AG-News, and \Cref{fig:al-curves-trec} for TREC dataset.

\begin{table}[]
\small
\centering
\begin{tabular}{l|rrr?r}
\toprule
 & $F_1^-$ & $F_1^0$ & $F_1^+$ & \\
\midrule
\textsc{subj} & $0$ & $1$ & $17$ & $18$  \\
\textsc{cola} & $3$ & $11$ & $4$ & $18$ \\
\textsc{agn} & $1$ & $4$ & $13$ & $18$ \\
\textsc{trec} & $0$ & $2$ & $16$ & $18$ \\
\midrule
R$\rightarrow$B & $0$ & $3$ & $9$ & $12$ \\
E$\rightarrow$B & $0$ & $4$ & $8$ & $12$ \\
B$\rightarrow$R & $1$ & $1$ & $10$ & $12$ \\
E$\rightarrow$R & $1$ & $2$ & $9$ & $12$ \\
B$\rightarrow$E & $1$ & $3$ & $8$ & $12$ \\
R$\rightarrow$E & $1$ & $5$ & $6$ & $12$ \\
\midrule
\textsc{ent} & $1$ & $5$ & $18$ & $24$ \\
\textsc{cs} & $2$ & $8$ & $14$ & $24$ \\
\textsc{ba} & $1$ & $5$ & $18$ & $24$ \\
\toprule
 & $4$ & $18$ & $50$ & \\
\bottomrule
\end{tabular}
\caption{Breakdown of datasets, acquisition$\to$consumer model pairs (denoted by initial letters), and AL methods by transferability measured via $F_1$ score at the end of the AL loop: negative ($-$), neutral ($0$), positive ($+$) transfer.}
\label{tab:groups-f1}
\end{table}

\begin{table*}[tb!]
\small
\centering
\begin{tabular}{l|rrr|rrr|rrr?r}
\toprule
& $\mathrm{AUC}^-$ & $\mathrm{AUC}^0$ & $\mathrm{AUC}^+$ & $\mathrm{AUC}_{10}^-$ & $\mathrm{AUC}_{10}^0$ & $\mathrm{AUC}_{10}^+$ & $F_1^-$ & $F_1^0$ & $F_1^+$ & \\
\midrule
\textsc{subj} & $0$ & $0$ & $9$ & $0$ & $0$ & $9$ & $0$ & $0$ & $9$ & $9$  \\
\textsc{cola} & $0$ & $3$ & $6$ & $1$ & $2$ & $6$ & $1$ & $4$ & $4$ & $9$ \\
\textsc{agn} & $1$ & $3$ & $5$ & $0$ & $0$ & $9$ & $0$ & $0$ & $9$ & $9$ \\
\textsc{trec} & $1$ & $3$ & $5$ & $0$ & $0$ & $9$ & $0$ & $0$ & $9$ & $9$ \\
\midrule
BERT & $1$ & $2$ & $9$ & $0$ & $0$ & $12$ & $0$ & $0$ & $12$ & $12$ \\
RoBERTa & $0$ & $4$ & $8$ & $1$ & $2$ & $9$ & $0$ & $3$ & $9$ & $12$ \\
ELECTRA & $1$ & $3$ & $8$ & $0$ & $0$ & $12$ & $1$ & $1$ & $10$ & $12$ \\
\midrule
\textsc{ent} & $1$ & $4$ & $7$ & $0$ & $1$ & $11$ & $0$ & $1$ & $11$ & $12$ \\
\textsc{cs} & $1$ & $4$ & $7$ & $1$ & $0$ & $11$ & $0$ & $2$ & $10$ & $12$ \\
\textsc{ba} & $0$ & $1$ & $11$ & $0$ & $1$ & $11$ & $1$ & $1$ & $10$ & $12$ \\
\toprule
 & $2$ & $9$ & $25$ & $1$ & $2$ & $33$ & $1$ & $4$ & $31$ & \\
\bottomrule
\end{tabular}
\caption{Breakdown of datasets, models, and AL methods in the groups based on the performance of regular AL: negative ($-$), neutral ($0$), positive ($+$) AL. }
\label{tab:groups-regular}
\end{table*}

\end{document}